\documentclass{elsart}

\usepackage{amssymb}
\usepackage{graphicx}
\usepackage{wrapfig}

\begin{document}

\begin{frontmatter}



\title{Locally Adaptive Block Thresholding Method with Continuity Constraint}


\author[1]{S. Hemachander},
\author[2]{A. Verma},
\author[3]{S. Arora},
\author[2]{Prasanta K. Panigrahi}\footnote[1]{E-mail: prasanta@prl.res.in}

\address[1]{University of Buffalo, NY, 14260, USA}
\address[2]{Physical Research Laboratory, Navrangpura, Ahmedabad, 380 009, India}
\address[3]{Dhirubhai Ambani Institute of Information and Communication Technology, Gandhinagar, 382 009, India}

\begin{abstract}
We present an algorithm that enables one to perform locally
adaptive block thresholding, while maintaining image continuity.
Images are divided into sub-images based some standard image
attributes and thresholding technique is employed over the
sub-images. The present algorithm makes use of the thresholds of
neighboring sub-images to calculate a range of values. The image
continuity is taken care by choosing the threshold of the
sub-image under consideration to lie within the above range. After
examining the average range values for various sub-image sizes of
a variety of images, it was found that the range of acceptable
threshold values is substantially high, justifying our assumption
of exploiting the freedom of range for bringing out local details.
\end{abstract}

\end{frontmatter}

\section{Introduction}

Applications like document image analysis [Dawoud and Kamel
(2001)], quality inspection of materials, non-destructive testing
[Sezgen and Sankur (2001)]  etc., require the concerned images to
be thresholded. Numerous methods to perform image thresholding
exist in the literature [Trier and Jain (1995), Sezgen and Sakur
(2004), Sahoo and Soltani (1998), Huang et al. (2005)].
Thresholding algorithms can be classified into the following main
categories: Histogram shape [Rosennfeld and Torre (1983), Sezan
(1985), Ramesh et al. (1995), Wang et al. (2002)], where the aim
of the algorithm is to find an optimal threshold that separates
two major peaks in the histogram, is implemented by sending a
smoothing filter on the histogram and then a difference filter or
by fitting the histogram with two Gaussians. But the main
disadvantage of histogram-based methods is their disregard of
spatial information. Image entropy based methods [Pun (1980),
Kapur et al.(1985), Li and Lee(1993),Li and Tam (1998)] use the
entropy of the image as a constraint for threshold selection. The
two common ways in which this can be done is by the maximization
of entropy of the thresholded image or minimization of cross
entropy between input image and the output binary image.

General image attributes [Tsai (1985), Hertz and Schafer (1998),
Gorman (1994), Arora et al. (2005)] can also be effectively used,
where the threshold is selected based on some similarity measure
between the original image and the binarized version of the image.
These can take the form of edges, shapes, color or other suitable
attributes like compactness or connectivity of the objects,
resulting from the binarization process or the coincidence of edge
fields. But the disadvantage of the above method lies in its
complexity and the relatively low image quality.

Clustering of gray level [Ridler and Calward (1978), Leung and Lam
(1996), Kittler and Illingworth (1986), Pal and Pal (1989)], based
methods aim to find two clusters in pixel distribution, a
foreground cluster and a background cluster. Various algorithms
exist for finding these clusters. Spatial information [Abutaleb
(1989)] utilize information of objects and background pixels in
the form of context probabilities, correlation functions,
co-occurrence probabilities, local linear dependence models of
pixels, two dimensional entropy etc.

Locally adaptive thresholding based methods [White and Rohrer
(1983), Niblack (1986), Lindquist (1999), Trier and Taxt (1995)],
are characterized by calculation of threshold at every pixel. The
value of the threshold depends upon some local parameters like
mean, variance, and surface fitting parameters or their suitable
combinations. In one approach, the gray value of the pixel is
compared with the average of the gray values in some neighborhood;
if the pixel is significantly larger than the average, it is
assigned as foreground, otherwise it is classified as background.
Another common method adapts the thresholding according to local
mean
\begin{math} (\mu) \end{math} and standard deviation
\begin{math} (\sigma) \end{math} over the window size. The threshold
at every pixel \begin{math}(i, j)\end{math} is calculated as
\begin{math} T (i, j)=\mu(i, j)+k.s (i, j)\end{math}, for a
suitable value of k. Niblack's method for thresholding is a
well-known example of this class. The calculation of threshold at
every pixel makes this technique relatively time consuming.

In approaches based on global thresholds, which are faster as
compared to their local counterparts, one calculates a single
threshold value for the entire image. A common example of this
class is Otsu's (1979) method of thresholding; this is an
iterative approach, which assumes that the gray level histogram is
the sum of two normal intensity distributions. Since the
thresholding is done once for the whole image, one may lose
certain local characteristics. Hence, the thresholding of images
based on local attributes have proved to be generally superior to
the global thresholding methods in terms of final image quality. A
number of the above thresholding methods suffer from the problem
of image continuity, which cannot be tolerated in applications
pertaining to medical imaging, remote sensing, optical character
recognition etc., where image continuity plays a crucial role.

The usual method of calculation of local threshold for every
pixel, with the help of information present in a window defined
around it, is computationally intensive. In this paper, we present
a hybrid method, where the threshold is calculated only once in a
window. This locally adaptive block thresholding (LABT) algorithm
makes use of the threshold values of the neighboring sub-images to
calculate a range. Image continuity is obtained by choosing the
threshold value of the sub-image under consideration to lie within
the range of values specified by the algorithm.

The above algorithm is applied to a wide variety of images and it
is observed that the local details are preserved to a great
extent. In addition to that, the algorithm fared better in terms
of time-complexity as compared to the other thresholding
techniques.

Before proceeding to the details of the technique, it is
convenient to define the following notations. In this text,
\begin{math}S_{m, n}\end{math} denotes a sub-image, where $\it m$ and $\it n$ denote the
position of the sub-image in the matrix of sub-images. Threshold
chosen for the sub-image \begin{math}S_{m, n}\end{math} on
application of an appropriate thresholding technique, is denoted
by \begin{math}OT_{m, n}\end{math} (Original Threshold). Threshold
value of sub-image \begin{math}S_{m, n}\end{math} after
application of the present algorithm, is denoted by
\begin{math}T_{m, n}\end{math}. The range of threshold values that
\begin{math}S_{m, n}\end{math} can take, without violating its
continuity with the upper (left) sub-image, is denoted by
\begin{math}UR_{m, n}~(LR_{m, n})\end{math}.

\section{Procedure For Local Block Thresholding With Continuity Constraint}
The given image needs to be divided into a number of sub-images of
size $\it m \times \it n$, where the values of $\it m$ and $\it n$
can be chosen on the basis of standard image attributes. In this
paper image variance has been chosen as that attribute. The image
having larger variance is divided into more number of sub-images
in order to bring out finer details, whereas an image with a lower
variance is divided into low number of sub-images in order to be
computationally inexpensive. The reason to divide a given image
into sub-images based on some attribute (image variance in this
case), is to balance between image quality and time complexity by
choosing the right sub-image size depending upon the application
under consideration and level of finer details to be extracted
from the image. In a number of cases e.g., optical character
recognition, the sub-image size is dictated by the image under
consideration. The sub-image size can also be left as a variable,
to be determined by the desired amount of image details. This may
have usefulness to medical imaging. The number of rows and columns
of the image are then converted to multiples of $\it m$ and $\it
n$ respectively.

Once the division of image has been done, the sub-images are
scanned from top-left to bottom-right. Constraint is then imposed
on the threshold selection of a sub-image by thresholds of upper
and left sub-images, i.e., continuity is sought between
\begin{math}S_{m, n}\end{math} and \begin{math}S_{m-1,
n}\end{math}, and \begin{math}S_{m, n}\end{math} and
\begin{math}S_{m, n-1}\end{math}.

Any threshold determination technique can be used to binarize the
sub-image \begin{math}S_{m, n}\end{math}, starting from
\begin{math}S_{1, 1}\end{math}. The thresholds \begin{math}T_{m-1, n}\end{math}
and \begin{math}T_{m, n-1}\end{math} of the neighboring sub-images
are used to impose constraint of continuity on the threshold
\begin{math}T_{m, n}\end{math} of the sub-image \begin{math}S_{m, n}\end{math}.
The choice of threshold \begin{math}T_{m, n}\end{math} of
\begin{math}S_{m, n}\end{math} is constrained to a range \begin{math}R_{m, n}\end{math}.
This range is determined using the threshold values of the
neighboring sub-images and the bordering pixel values of the
sub-image under consideration. Any value in the range
\begin{math}R_{m, n}\end{math} when used to threshold the columns
(rows) of \begin{math}S_{m, n}\end{math}, that borders the
adjacent sub-image \begin{math}S_{m-1, n}~(S_{m, n-1})\end{math},
classify them into foreground or background, in the same way
\begin{math}T_{m-1, n}~(T_{m, n-1})\end{math} classifies the pixels of those borders. Stating in
symbolic terms, if \begin{math}\tau T_{m, n}\end{math} denotes
thresholding operation, i.e., classifying every pixel of a given
array/matrix into foreground or background, using
\begin{math}T_{m, n}\end{math}, the constraint is then stated as:

\begin{equation}
\tau T_{m,n}(S_{m,n}(outer~lining))=\tau
T_{m-1,n}(S_{m,n}(outer~lining))
\end{equation}
if the outer lining is a row of pixels and,

\begin{equation}
\tau T_{m,n}(S_{m,n}(outer~lining))=\tau
T_{m,n-1}(S_{m,n}(outer~lining))
\end{equation}
if the outer lining is a column of pixels

The range of threshold values \begin{math}R_{m, n}\end{math}, that
a sub-image can take while maintaining image continuity with the
upper and side block, is determined in the following manner. An
array comprising of threshold Tm-1,n of the upper sub-image and
pixel values of \begin{math}R_{m, n}\end{math}'s uppermost row
(say, $\it X$), which borders the upper sub-image with maximum and
minimum pixel values, is created. Pixel values in $\it X$, which
are equal to \begin{math}T_{m, n}\end{math}, are deleted before
$\it X$ is added to the array in order get values other than
\begin{math}T_{m-1, n}\end{math}. Appending minimum and maximum
pixel values ensures the presence of values, greater and less than
\begin{math}T_{m-1, n}\end{math}. Let the values that are immediately lower and greater than
\begin{math}T_{m-1, n}\end{math} in the array, be
\begin{math}R_{l1}\end{math} and \begin{math}R_{h1}\end{math}, respectively. Then the range
dictated by the upper sub-image is

\begin{displaymath}
UR_{m, n}= [R_{l1+1},R_{h1}].
\end{displaymath}

The classification of foreground and background pixels is done,
assuming the definition of thresholding as \begin{math} X
<\end{math} threshold \begin{math}=> X = \end{math} background,
and \begin{math}X \geq\end{math} threshold \begin{math}=> X =
\end{math} foreground.

The same procedure is applied to determine the range
\begin{math}LR_{m, n}\end{math} dictated by the left sub-image.
Here, $\it X$ is the column of \begin{math}S_{m, n}\end{math}
which borders the left sub-image and the threshold to be added to
the array is \begin{math}T_{m, n-1}\end{math} .The effective
range, within which the threshold of \begin{math}S_{m,
n}\end{math}, has to be selected to avoid discontinuity, is

\begin{displaymath}
R_{m, n}= UR_{m,n}\cap LR_{m,n}.
\end{displaymath}

\section{Algorithm}

The salient features of the proposed LABT algorithm can be stated
in the following manner:

1)  The image is divided into number of same-sized rectangular
sub-images based on the variance of the whole image. Other
attributes of the image can also be used for this purpose.

2) Image is then made into a multiple of the sub-image, by a
suitable operation.

3)  Starting from $S_{1,1}$, operations are performed on the
sub-images row-wise, i.e., the image is scanned from top-left to
bottom right.

4)  An original threshold $OT_{m,n}$ of $S_{m,n}$, is determined,
using a suitable thresholding technique.

5)  The range $R_{m,n}$ for the sub-window under consideration is
worked out, using $T_{m-1,n}$ and $T_{m,n-1}$ with the help of
above method.

6)  For $S_{1,1}$, the threshold $T_{1,1}=OT_{1,1}$. For the
sub-images in the topmost row (leftmost column), continuity is
maintained only with the left (upper) sub-images.

7)  In case $OT_{m,n}$ falls out of $R_{m,n}$, it is brought to
the nearest extreme of $R_{m,n}$, using the above specified
procedure, and denoted by $T_{m,n}$.

The above algorithm thus ensures that continuity is maintained
across sub-images. Sub-image size can be changed depending upon
the purpose, i.e., a smaller sub-image size can be taken to bring
out finer details, whenever it is necessary. Bigger size
sub-images are advisable for document image thresholding, where
fuzzy outlines of letters need to be made well defined. A bigger
sub-image size will help in keeping the threshold almost constant,
across letters, thereby providing a consistent cut-off for
removing fuzziness.

\section{Results and Observations}

\begin{figure}
\resizebox{1.75in}{1.75in}{\includegraphics{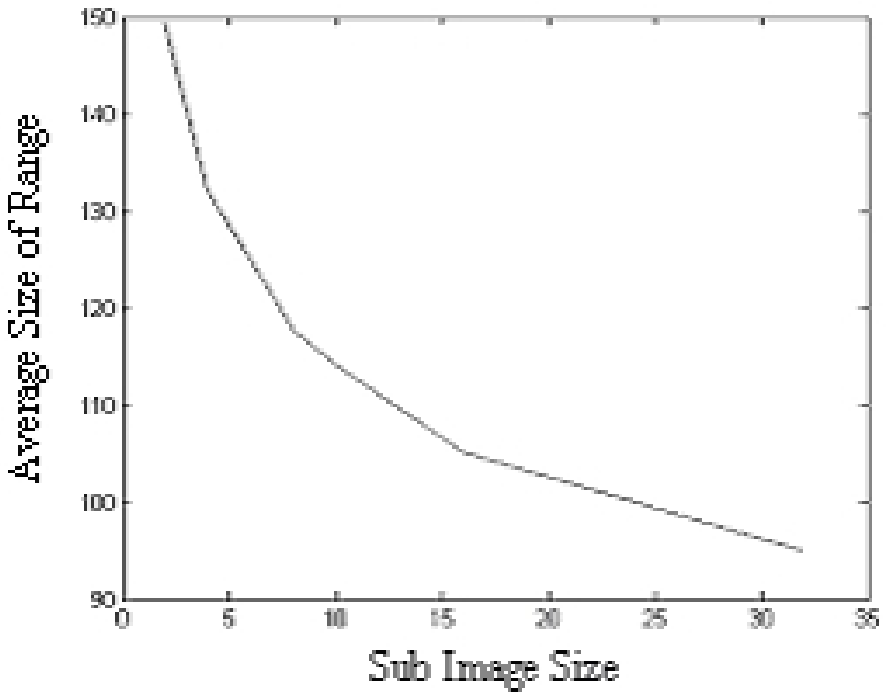}}
\resizebox{1.75in}{1.75in}{\includegraphics{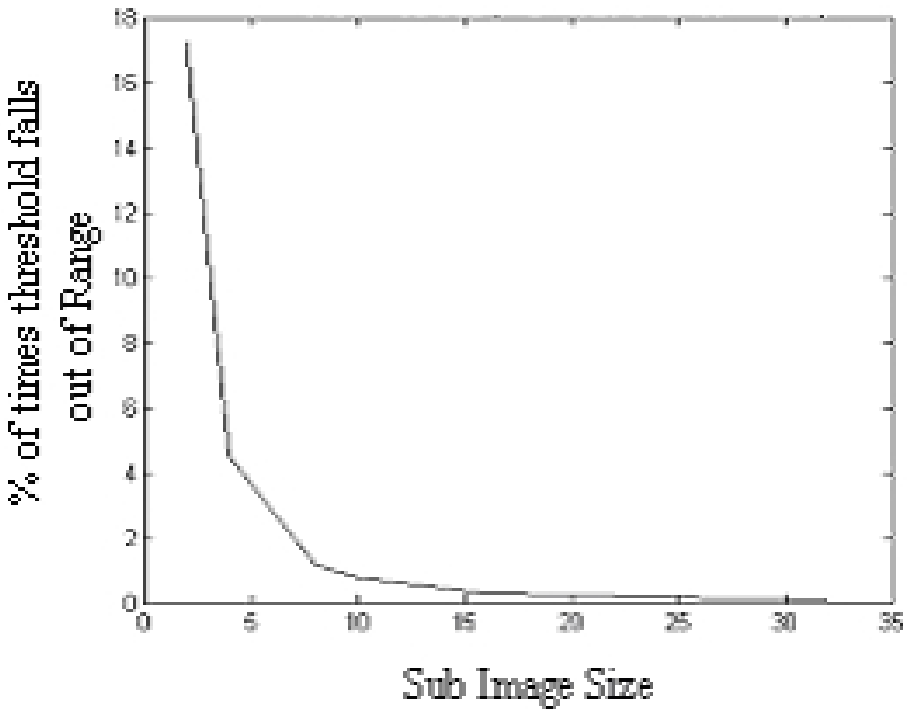}}
\resizebox{1.75in}{1.75in}{\includegraphics{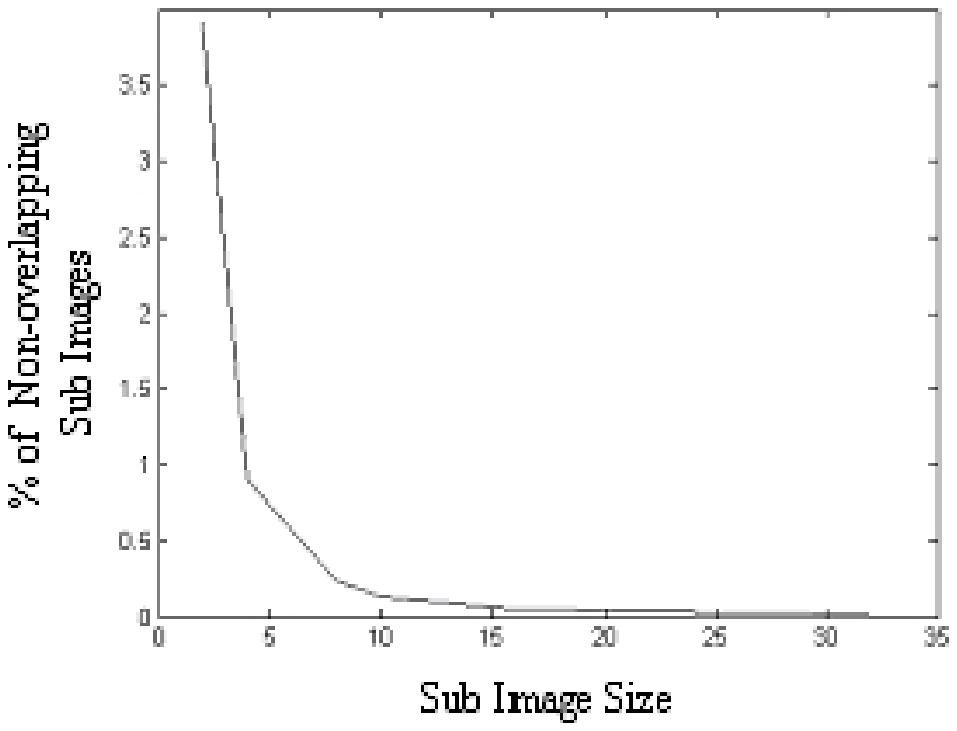}}
\begin{tabbing}
\hspace{.88in}\small{(a)\hspace{1.6in}(b)\hspace{1.6in}(c)}\\
\end{tabbing}
\caption{Plots showing dependence of three important
characteristics on sub image size (averaged over 35 images)}
\end{figure}

We observed reduction in the average size of the range $R_{m,n}$
with increasing sub-image size. This can be seen from Fig. 1a.
This reduction in size is because of the availability of elements
nearer to $T_{m,n}$ in the bordering array $\it X$, when the
sub-image size gets bigger. Since the algorithm starts from
$S_{1,1}$ and propagates downwards, it is preferable to binarize
$S_{1,1}$ with threshold obtained by applying threshold over the
whole image.

\begin{figure}
\resizebox{1.75in}{1.75in}{\includegraphics{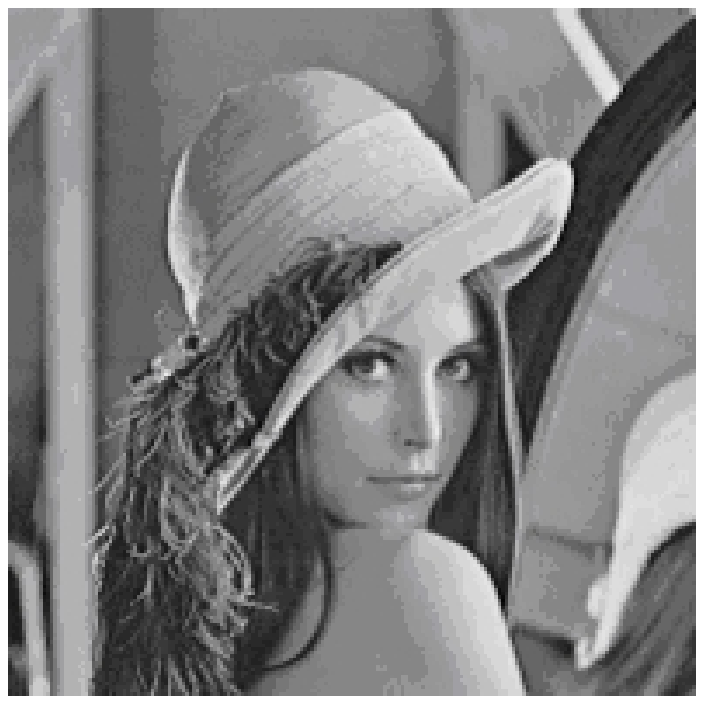}}
\resizebox{1.75in}{1.75in}{\includegraphics{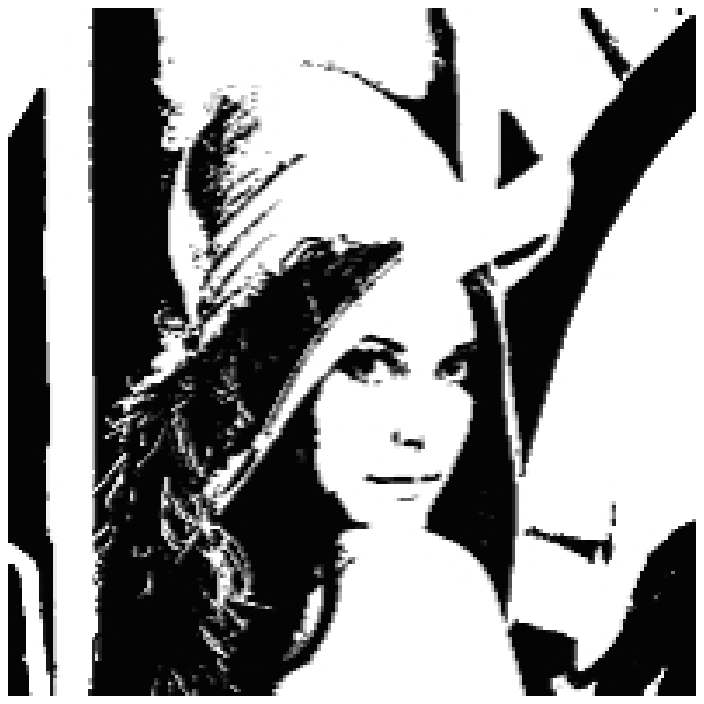}}
\resizebox{1.75in}{1.75in}{\includegraphics{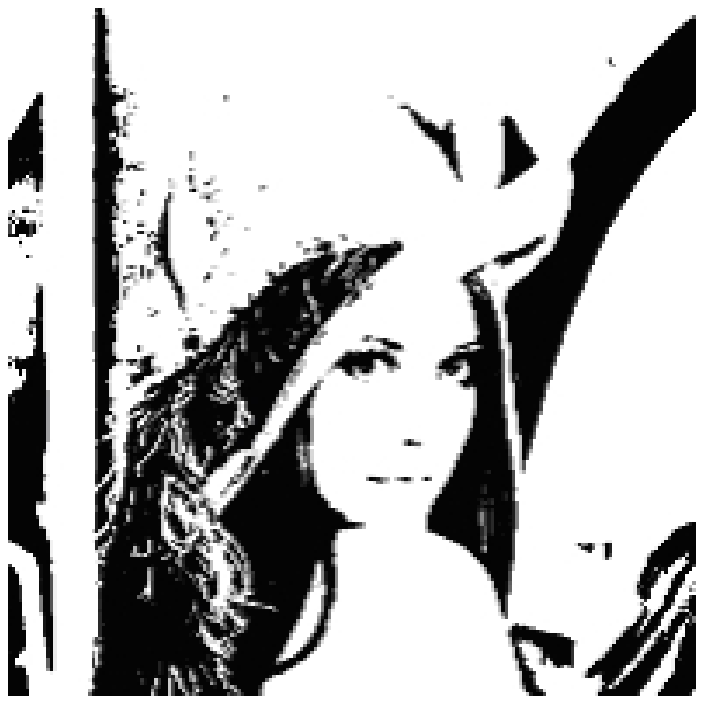}}
\begin{tabbing}
\hspace{.88in}\small{(a)\hspace{1.5in}(b)\hspace{1.5in}(c)}\\
\end{tabbing}
\resizebox{1.75in}{1.75in}{\includegraphics{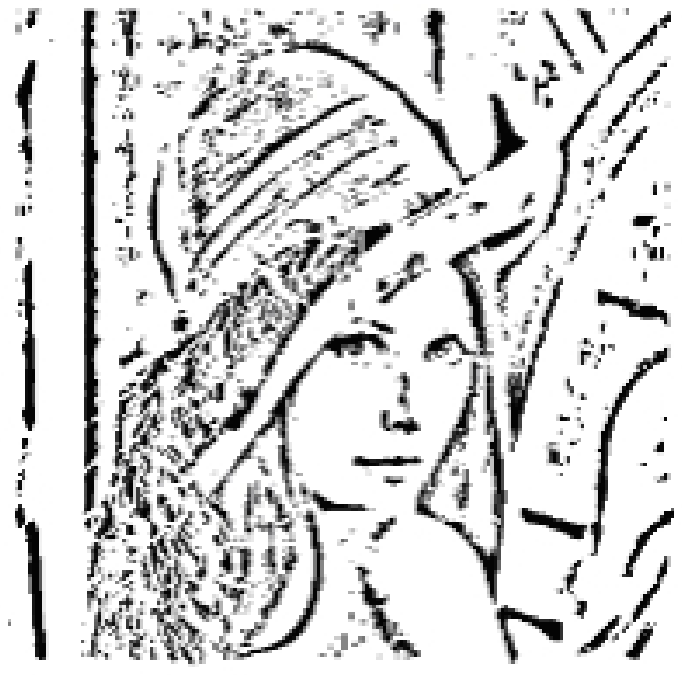}}
\resizebox{1.75in}{1.75in}{\includegraphics{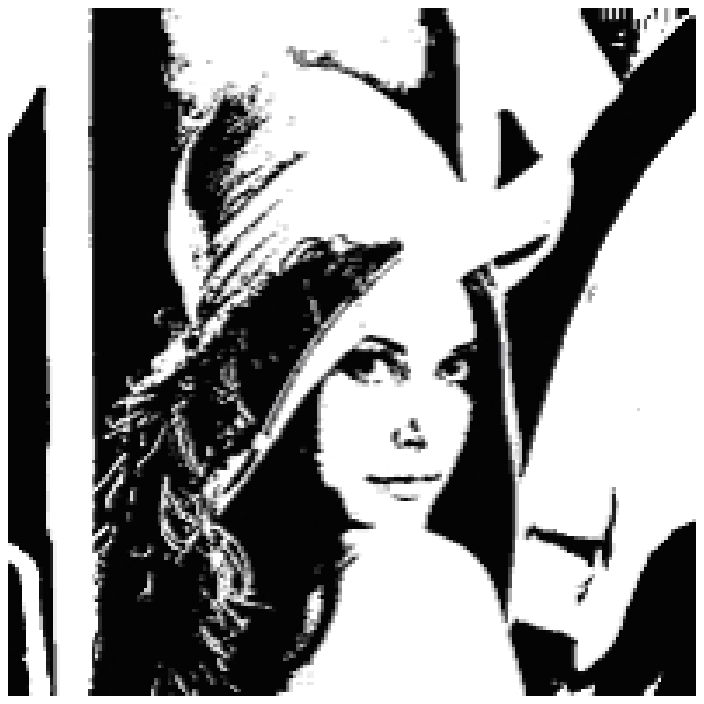}}
\resizebox{1.75in}{1.75in}{\includegraphics{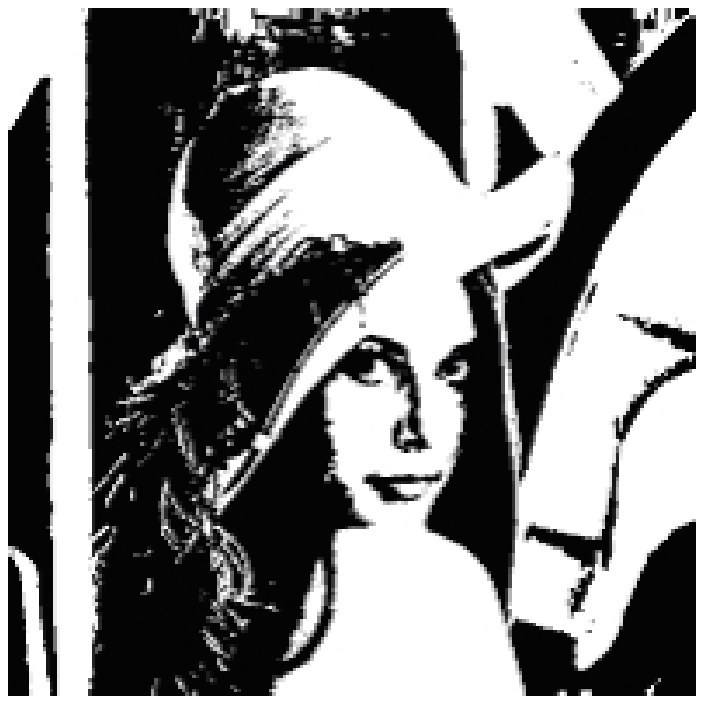}}
\begin{tabbing}
\hspace{.88in}\small{(d)\hspace{1.5in}(e)\hspace{1.5in}(f)}
\end{tabbing}
\caption{Comparison between the results of various methods, (a)
Original  image, (b) Thresholded using ADCDF, (c) Thresholded
using Otsu, (d) Thresholded using Niblack, (e) Thresholded using
ADCDF with LABT, (f) Thresholded using Otsu with LABT.}
\end{figure}

The second plot (Fig. 1b) shows the variation of the fraction of
times threshold exceeds the range constraint; with sub-image size
averaged over 35 images. The fraction of times $OT_{m,n}$ falling
outside $R_{m,n}$ decreases with increase in sub-image size. This
is due to stabilization of threshold across sub-images, when the
sub-image size is increased. This is as expected, given the
large-scale homogeneity present in numerous images.

\begin{figure}
\begin{centering}
\resizebox{2.2in}{2.2in}{\includegraphics{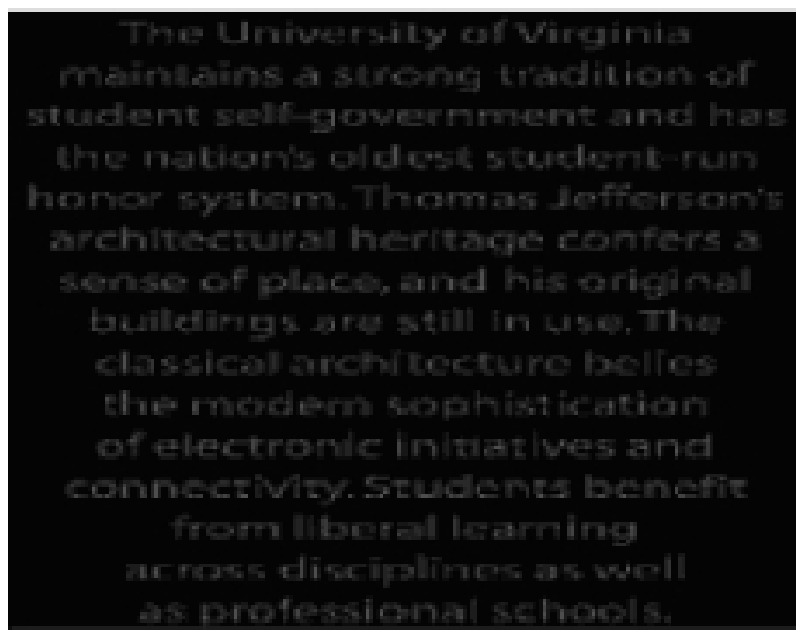}}
\resizebox{2.2in}{2.2in}{\includegraphics{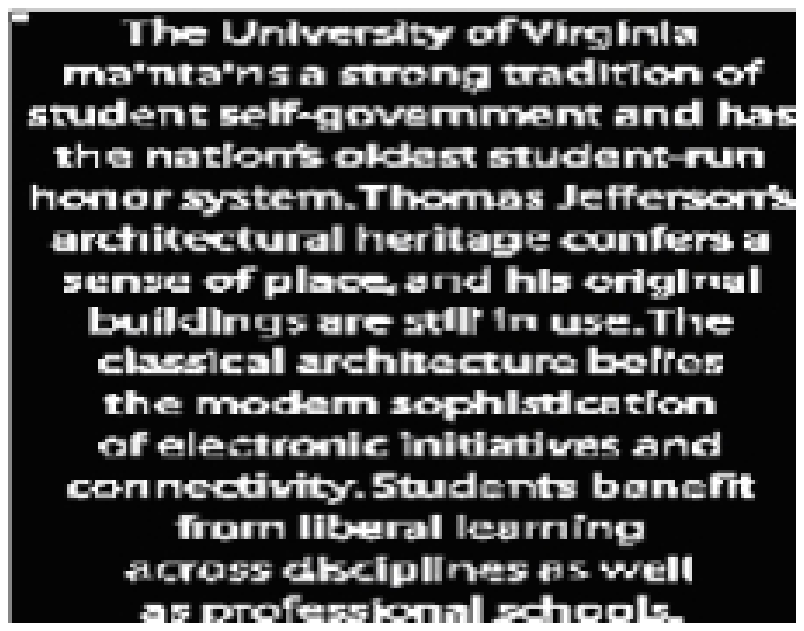}}
\begin{tabbing}
\hspace{1.5in}\small{(a)\hspace{2.2in}(b)}\\
\end{tabbing}
\caption{(a) Original text image, (b) Thresholded using LABT.}
\end{centering}
\end{figure}

We are presenting a few images for which the familiar method of
thresholding, i.e., area division of cumulative distribution
function (ADCDF), is used to binarize the sub-images. We have also
used Otsu's algorithm for the same purpose. Superior thresholding
methods, when used in conjunction with this algorithm, will give
far better results. For the purpose of illustrating the efficacy
of our procedure and comparison, we have also presented the
binarized images, using Otsu (global) and Niblack (local)
thresholding methods in Fig. 2. One clearly sees that the present
locally adaptive block thresholding method clearly does well in
terms of extracting local features as well as retaining the visual
image quality. We have checked this property of LABT in a variety
of images.

\begin{figure}
\centering
\resizebox{6in}{4.5in}{\includegraphics{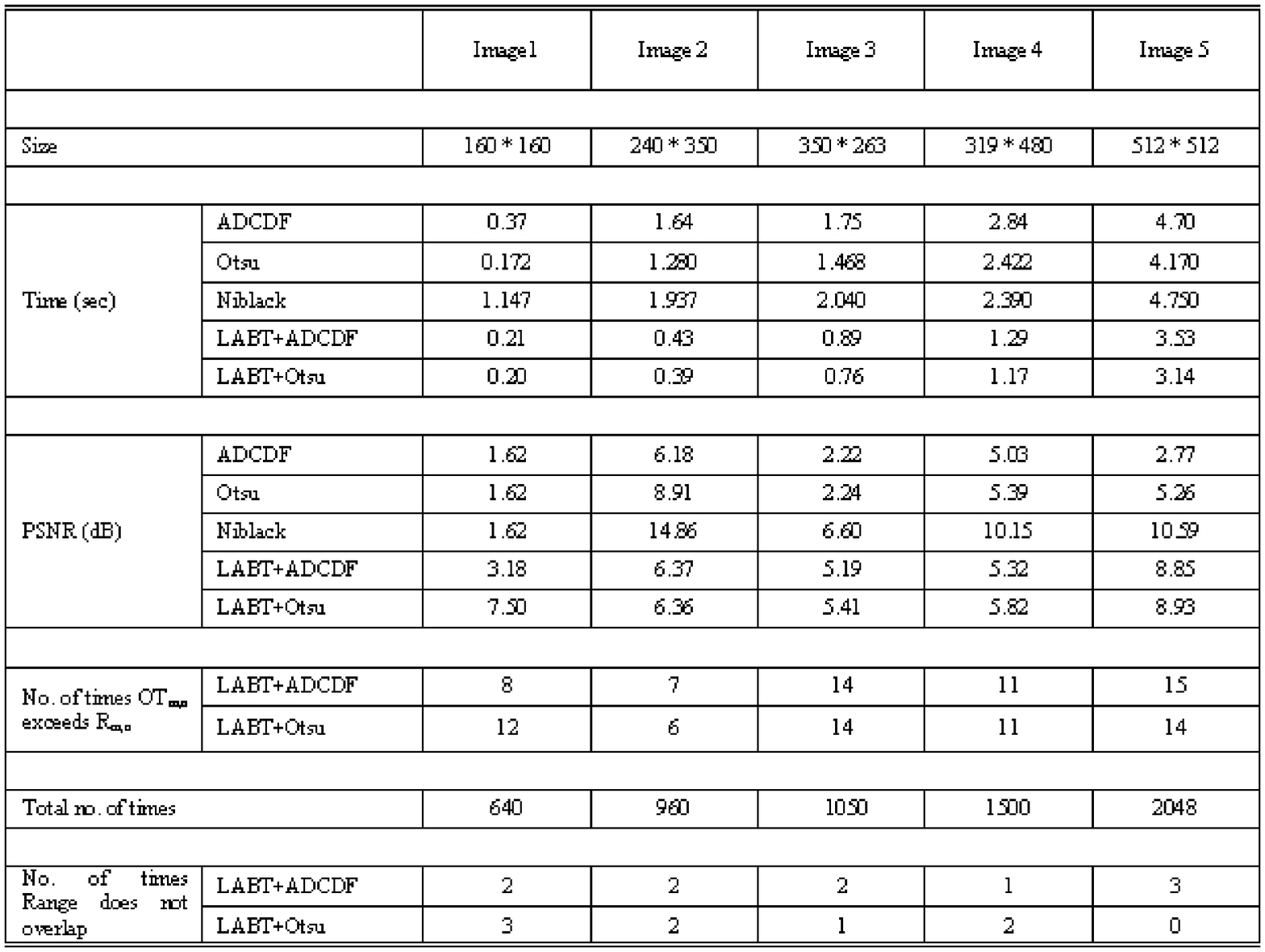}}
\small{\newline\newline Table 1 \\ Quantitative comparision of
various thresholding methods}
\end{figure}

\begin{figure}
\begin{centering}
\resizebox{1.75in}{1.75in}{\includegraphics{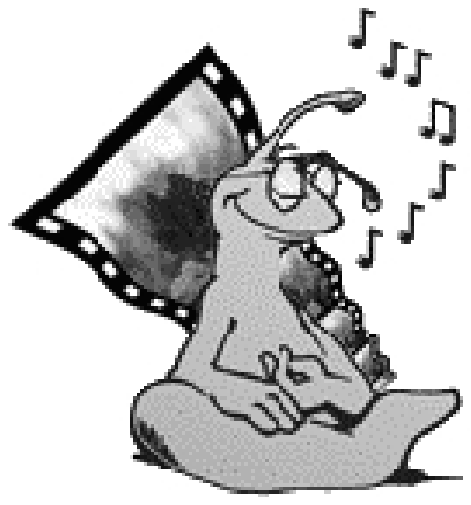}}
\resizebox{1.75in}{1.75in}{\includegraphics{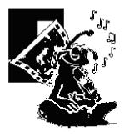}}
\begin{tabbing}
\hspace{1.88in}\small{(a)\hspace{1.5in}(b)}\\
\end{tabbing}
\resizebox{1.75in}{1.75in}{\includegraphics{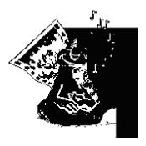}}
\resizebox{1.75in}{1.75in}{\includegraphics{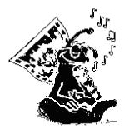}}
\begin{tabbing}
\hspace{1.88in}\small{(c)\hspace{1.5in}(d)}
\end{tabbing}
\caption{(a) Original image, (b) Inverted threshold, (c)
Thresholded image, and (d) Final image obtained after ORing.}
\end{centering}
\end{figure}

For text images, thresholding followed by morphological operation
like thinning gives good results (Fig. 3a \& Fig 3b). It is
advisable to choose the sub-image size to be more than the average
object size in the image. This ensures the whole object, to be
uniformly classified as background or foreground, and avoids
classification of within-object variation.

The computational time and PSNR for different thresholding
techniques, implemented with and without the locally adaptive
block thresholding (LABT), has been shown in table 1. It is quite
obvious from the results that the standard thresholding techniques
fare better when applied in conjunction with our algorithm. The
table also shows that the number of times threshold exceeds the
range, and number of times the range dictated by upper and side
sub-image does not overlap for three different thresholding
algorithms are quite small. This justifies our assumption of
exploiting the freedom of range for bringing out local details.

To avoid possible errors arising from the scanning of the image,
row-wise from top to bottom, one can scan the image in different
ways and perform an ORing operation of the different images, as
specified below:

1)  Image is thresholded in the usual way.

2) Invert the original image upside down and threshold the image.
Then invert it back to the original state.

3)  Invert the given image right side left and threshold the
image. Then invert it back.

4)  ORing operation is carried out on the above images to get the
resulting image, which is equivalent to scanning the image in
different ways and ORing them. Not just scanning row-wise from top
to bottom.

The results of the ORing operation thus give superior results as
shown in Fig. 4. One can see much clearer local details in the
final image.

\section{Conclusion and Discussion}
In this paper, a new locally adaptive block thresholding method
has been proposed, which acts as a hybrid between known local and
global methods. It can also be used in conjunction with other
methods of binarization to bring out details of an image. It
should be emphasized that the same is accomplished without
introducing too much of time complexity, an extremely desired
attribute of any binarization scheme. The present algorithm has
been designed to ensure that the transitions between sub-windows
are maintained continuously. This maintains image continuity.

The efficacy of the method has been demonstrated in the context of
a variety of images of different types. This procedure may also be
useful when a variable window size is required. The portions of an
image requiring detailed investigations may be divided into finer
sub images, whereas other portions of the image can be divided
into bigger box sizes. In this case, one needs to explore the
problem of boundary mismatch and continuity more carefully. The
boundary mismatch can be possibly taken care by pushing the
boundary of the block that created the mismatch, till the selected
threshold falls within the range. This problem is currently under
investigation and will be reported elsewhere.

\end{document}